\pdfoutput=1

\documentclass[11pt]{article}

\usepackage[]{acl}

\usepackage{times}
\usepackage{latexsym}
\usepackage{multicol}
\usepackage[T1]{fontenc}

\usepackage[utf8]{inputenc}

\usepackage{microtype}
\usepackage{xspace}
\usepackage{booktabs}
\usepackage{multirow}
\usepackage{makecell}
\usepackage{adjustbox}
\usepackage{bm}
\usepackage{amsmath}
\usepackage{balance}
\usepackage{enumitem}
\usepackage{xcolor,colortbl}
\usepackage{arydshln}
\usepackage{pifont}
\usepackage{float}
\newcommand{\cmark}{\ding{51}}%
\newcommand{\xmark}{\ding{55}}%

\title{\ours: Patching Visual Modality to Textual-Established \\Information Extraction}

\newcommand{\ours}{PV2TEA\xspace}

\usepackage{lipsum}                     
\usepackage{xargs}                      
\usepackage[colorinlistoftodos,prependcaption,textsize=tiny]{todonotes}
\newcommandx{\cmt}[2][1=]{\todo[linecolor=yellow,backgroundcolor=yellow!25,bordercolor=yellow,#1]{#2}}

\author{Hejie Cui$^1\thanks{~\hspace{0.03in}Work was done when Hejie was an intern at Amazon.}$ , Rongmei Lin$^2$, Nasser Zalmout$^2$, Chenwei Zhang$^2$, \\ \bf Jingbo Shang$^{3}$,  Carl Yang$^1$, Xian Li$^2$ \\
$^1$ Emory University, GA, USA \\ $^2$ Amazon.com Inc, WA, USA \\ $^3$ University of California, San Diego, CA, USA \\
\texttt{\{hejie.cui, j.carlyang\}@emory.edu}, \texttt{jshang@ucsd.edu} \\
\texttt{\{linrongm, nzalmout, cwzhang, xianlee\}@amazon.com}
}

\begin{document}
\maketitle
\begin{abstract}
Information extraction, e.g., attribute value extraction, has been extensively studied and formulated based only on text. 
However, many attributes can benefit from image-based extraction, like color, shape, pattern, among others. The visual modality has long been underutilized, mainly due to multimodal annotation difficulty. In this paper, we aim to patch the visual modality to the textual-established attribute information extractor. The cross-modality integration faces several unique challenges: (C1) images and textual descriptions are loosely paired intra-sample and inter-samples; (C2) images usually contain rich backgrounds that can mislead the prediction; (C3) weakly supervised labels from textual-established extractors are biased for multimodal training. We present \ours, an encoder-decoder architecture equipped with three bias reduction schemes: (S1) Augmented label-smoothed contrast to improve the cross-modality alignment for loosely-paired image and text; (S2) Attention-pruning that adaptively distinguishes the visual foreground; (S3) Two-level neighborhood regularization that mitigates the label textual bias via reliability estimation. Empirical results on real-world e-Commerce datasets\footnote{The code and the human-annotated datasets with fine-grained source modality labels of gold values are available at \url{https://github.com/HennyJie/PV2TEA}. 
} demonstrate up to 11.74\% absolute (20.97\% relatively) $\text{F}_1$ increase over unimodal baselines.
\end{abstract}

\section{Introduction}
\label{sec:intro}
\begin{figure}
    \centering
    \includegraphics[width=\linewidth]{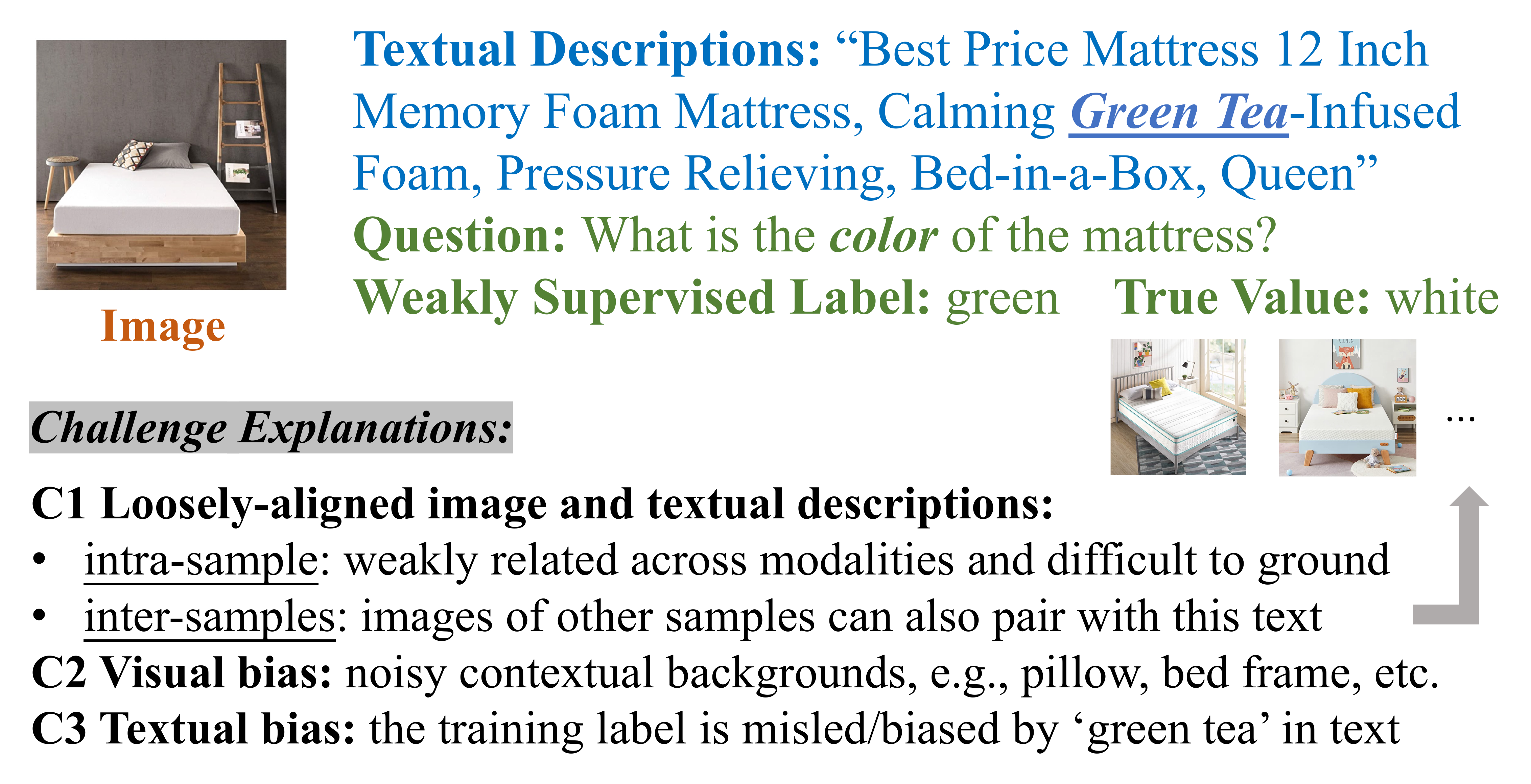}
    \caption{Illustration of multimodal attribute extraction and the challenges in cross-modality integration.} 
    \label{fig:intro-task-challenge}
\end{figure}
Information extraction, e.g., attribute value extraction, aims to extract structured knowledge triples, i.e., (\textit{sample\_id, attribute, value}), from the unstructured information. 
As shown in Figure \ref{fig:intro-task-challenge}, the inputs include text descriptions and images (optional) along with the queried attribute, and the output is the extracted value.
In practice, textual description has played as the main or only input in mainstream approaches for automatic attribute value extraction \cite{zheng2018opentag, xu2019scaling, wang2020learning, DBLP:conf/acl/KaramanolakisMD20, DBLP:conf/acl/YanZLGRD20, ding2022ask}. Such models perform well when the prediction targets are inferrable from the text. 

As the datasets evolve, interest in incorporating visual modality naturally arises, especially for image-driven attributes, e.g., \textit{Color}, \textit{Pattern}, \textit{Item Shape}. Such extraction tasks rely heavily on visual information to obtain the correct attribute values. The complementary information contained in the images can improve recall in cases where the target values are not mentioned in the texts. In the meantime, the cross-modality information can help with ambiguous cases and improve precision. 

However, extending a single-modality task to multi-modality can be very challenging, especially due to the lack of annotations in the new modality. Performing accurate labeling based on multiple modalities requires the annotator to refer to multiple information resources, leading to a high cost of human labor. Although there are some initial explorations on multimodal attribute value extraction~\citep{DBLP:conf/emnlp/ZhuWLWHZ20,lin2021pam,de2022multi}, 
all of them are fully supervised and overlook the resource-constrained setting of building a multimodal attribute extraction framework based on the previous textual-established models.
In this paper, we aim to patch the visual modality to attribute value extraction by leveraging textual-based models for weak supervision, thus reducing the manual labeling effort. 


\noindent\textbf{Challenges.} 
Several unique challenges exist in visual modality patching: 
\textbf{C1.} Images and their textual descriptions are usually \textit{loosely aligned} in two aspects: From the \underline{intra-sample} aspect, they are usually weakly related considering the rich characteristics, making it difficult to ground the language fragments to the corresponding image regions; From the \underline{inter-samples} aspect, it is commonly observed that the text description of one sample may also partially match the image of another. As illustrated in Figure \ref{fig:intro-task-challenge}, the textual description of the \textit{mattress} product is fragmented and can also correspond to other images in the training data. Therefore, traditional training objectives for multimodal learning such as binary matching~\citep{kim2021vilt} or contrastive loss~\citep{radford2021learning} that only treat the text and image of the same sample as positive pairs may not be appropriate. 
\textbf{C2.} Bias can be brought by the \textit{visual input} from the \textit{noisy contextual background}. The images usually not only contain the interested object itself but also demonstrate a complex background scene. 
Although the backgrounds are helpful for scene understanding, they may also introduce spurious correlation in a fine-grained task such as attribute value extraction, which leads to imprecise prediction~\citep{DBLP:conf/iclr/XiaoEIM21,kan2021zero}. 
\textbf{C3.} Bias also exists in \textit{language perspective} regarding the \textit{biased weak labels} from textual-based models. As illustrated in Figure \ref{fig:intro-task-challenge}, the color label of \textit{mattress} is misled by `\textit{green tea infused}' in the text. 
These noisy labels can be more catastrophic for a multimodal model due to their incorrect grounding in images.
Directly training the model with these biased labels can lead to gaps between the stronger language modality and the weaker vision modality~\citep{yufine}.


\noindent\textbf{Solutions.}
We propose \ours, a sequence-to-sequence backbone composed of three modules: visual encoding, cross-modality fusion and grounding, and attribute value generation, each with a bias-reduction scheme dedicated to the above challenges: \textbf{S1.} To better integrate the \textit{loosely-aligned texts and images}, we design an augmented label-smoothed contrast schema for cross-modality fusion and grounding, which considers both the intra-sample weak correlation and the inter-sample potential alignment, encouraging knowledge transfer from the strong textual modality to the weak visual one. \textbf{S2.} During the visual encoding, we equip \ours with an attention-pruning mechanism that adaptively distinguishes the distracting background and \textit{attends to the most relevant regions} given the entire input image, aiming to improve precision in the fine-grained task of attribute extraction. \textbf{S3.} To mitigate the bias from \textit{textual-biased weak labels}, a two-level neighborhood regularization based on visual features and previous predictions, is designed to emphasize trustworthy training samples while mitigating the influence of textual-biased labels. In this way, the model learns to generate more balanced results rather than being dominated by one modality of information. 
In summary, the main contributions of \ours are three-fold:
\begin{itemize}[nosep,leftmargin=*]
    \item We propose \ours, an encoder-decoder framework effectively patching up visual modality to textual-established attribute value extraction. 
    \item We identify three unique challenges in patching visual modality for information extraction, with solutions for \textit{intra-sample and inter-samples loose alignment} and bias from \textit{complex visual background} and \textit{textual-biased labels}. 
    \item We release three human-annotated datasets with modality source labels of the gold values to facilitate fine-grained evaluation. Extensive results validate the effectiveness of \ours. 
\end{itemize}
\section{Preliminaries}
\label{sec:prelim}
\subsection{Problem Definition}
We consider the task of automatic attribute extraction from multimodal input, i.e., textual descriptions and images. 
Formally, the input is a query attribute $\mathcal{R}$ and a text-image pairs dataset $\mathcal{D} = \{\mathcal{X}_n\}_{n=1}^N = \{(\mathcal{I}_n, \mathcal{T}_n, c_n)\}_{n=1}^N$ consisting of $N$ samples (e.g., products), where $\mathcal{I}_n$ represents the profile image of $\mathcal{X}_n$, 
$\mathcal{T}_n$ represents the textual description and $c_n $ is the sample category (e.g., product type). The model is expected to infer attribute value $y_n$ of the query attribute $\mathcal{R}$ for sample $\mathcal{X}_n$. We consider the challenging setting with open-vocabulary attributes, where the number of candidate values is extensive and $y_n$ can contain either single or multiple values. 

\subsection{Motivating Analysis on the Textual Bias of Attribute Information Extraction}
\label{sec:motivation}
Existing textual-based models or multimodal models directly trained with weak labels suffer from a strong bias toward the texts. As illustrated in Figure \ref{fig:intro-task-challenge}, the training label for the \textit{color} attribute of the \textit{mattress} is misled by `\textit{green tea infused}' from the textual profile. Models trained with such textual-shifted labels will result in a learning ability gap between modalities, where the model learns better from the textual than the visual modality. 
\begin{figure}
    \centering
    \includegraphics[width=\linewidth]{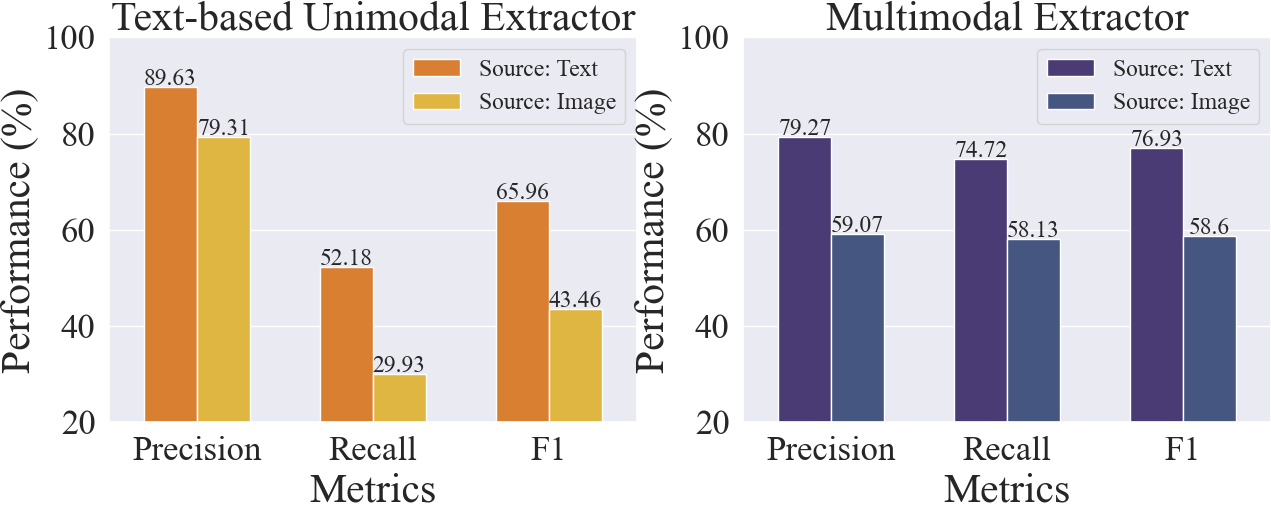}
    \caption{Source-aware evaluation of existing unimodal and multimodal models on the textual-biased issue.}
    \label{fig:source_aware_eva}
\end{figure}
To quantitatively study the learning bias, we conduct fine-grained source-aware evaluations on a real-world e-Commerce dataset with representative unimodal and multimodal methods, namely OpenTag~\citep{zheng2018opentag} with the classification setup and PAM~\citep{lin2021pam}. Specifically, for each sample in the test set, we collect the source of the gold value (i.e., text or image). Experiment results are shown in Figure \ref{fig:source_aware_eva}, where label \textit{Source: Text} indicates the gold value is present in the text, while label \textit{Source: Image} indicates the gold value is absent from the text and must be inferred from the image. 
It is shown that both the text-based unimodal extractor and multimodal extractor achieve impressive results when the gold value is contained in the text. 
However, when the gold value is not contained in the text and must be derived from visual input, the performance of all three metrics drops dramatically, indicating a strong textual bias and dependence of existing models.

\section{\ours}
\label{sec:method}
\begin{figure*}
    \centering
    \includegraphics[width=\linewidth]{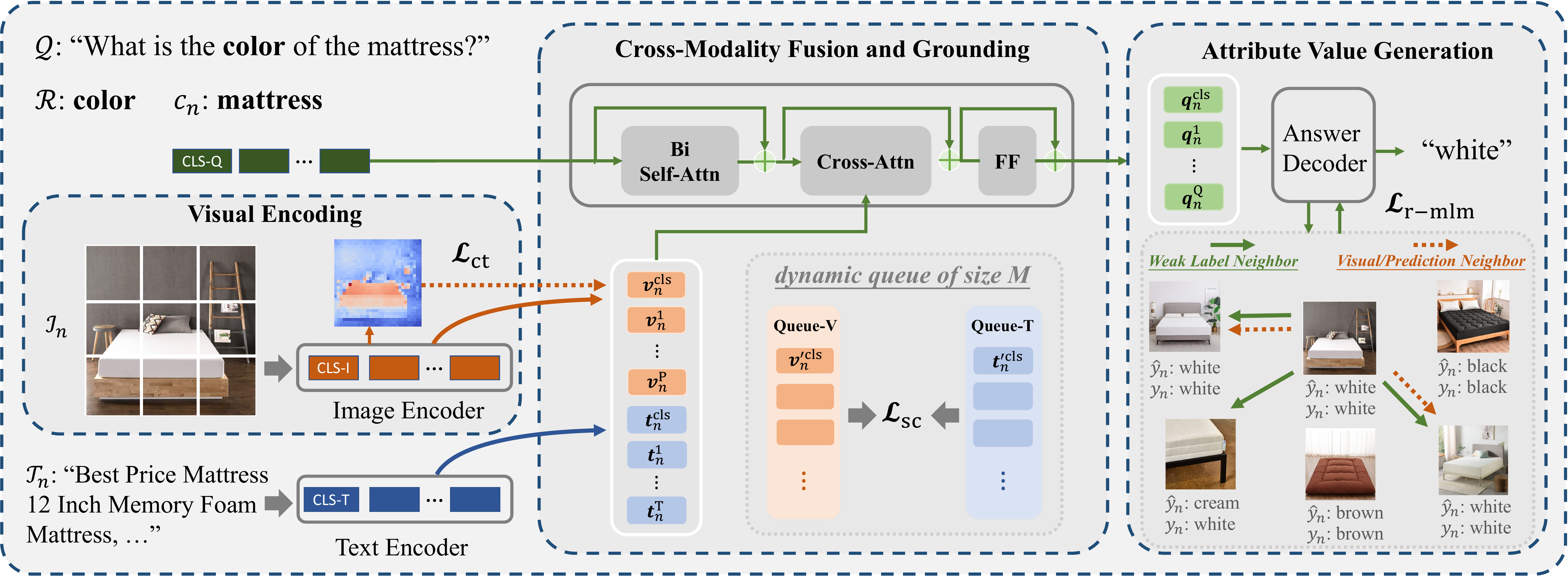}
    \caption{The overview of \ours model architecture with three modules, where each of them is equipped with a bias reduction scheme corresponding to the discussed challenges in Figure \ref{fig:intro-task-challenge}.}
    \label{fig:framework}
\end{figure*}
We present the backbone architecture and three bias reduction designs of \ours, shown in Figure \ref{fig:framework}. The backbone is formulated based on visual question answering (VQA) composed of three modules:  \\
\noindent{(1) \textbf{Visual Encoding.}}
We adopt the Vision Transformer (ViT) \citep{DBLP:conf/iclr/DosovitskiyB0WZ21} as the visual encoder. The given image $\mathcal{I}_n$ is divided into patches and featured as a sequence of tokens, with a special token \verb|[CLS-I]|appended at the head of the sequence, whose representation $\bm v_n^\text{cls}$ stands for the whole input image $\mathcal{I}_n$. \\
\noindent{(2) \textbf{Cross-Modality Fusion and Grounding.}}
Following the VQA paradigm, we define the question prompt as ``\textit{What is the $\mathcal{R}$ of the $c_n$}?", with a special token \verb|[CLS-Q]| appended at the beginning. A unimodal BERT \citep{DBLP:conf/naacl/DevlinCLT19} encoder is adopted to produce token-wise textual representation from sample profiles (title, bullets, and descriptions). The visual representations of $\text{P}$ image patches $\bm v_n=[\bm v_n^{\text{cls}}, \bm v_n^{1}, \ldots, \bm v_n^\text{P}]$ are concatenated with the textual representation of $\text{T}$ tokens $\bm t_n=[\bm t_n^{\text{cls}}, \bm t_n^{1}, \ldots, \bm t_n^\text{T}]$, which is further used to perform cross-modality fusion and grounding with the question prompt through cross-attention. The output $\bm q_n=[\bm q_n^{\text{cls}}, \bm q_n^{1}, \ldots, \bm q_n^\text{Q}]$ is then used as the grounded representation for the answer decoder. \\
\noindent{(3) \textbf{Attribute Value Generation.}}
We follow the design from \citep{DBLP:conf/icml/0001LXH22}, 
where each block of the decoder is composed of a causal self-attention layer, a cross-attention layer, and a feed-forward network. The decoder takes the grounded multimodal representation as input and predicts the attribute value $\hat{y}_n$ in a generative manner\footnote{We compared the settings of generation and classification for the attribute value extractor. See results in Section \ref{sec:cla-gen}.}. \\
\noindent{\textbf{Training Objectives.}} 
The overall training objective of \ours is formulated as
\begin{equation}
\small
\setlength{\abovedisplayskip}{3pt}
\setlength{\belowdisplayskip}{3pt}
\mathcal{L}=\mathcal{L}_{\text{sc}}+\mathcal{L}_{\text{ct}}+\mathcal{L}_{\text{r-mlm}},
\end{equation}
where the three loss terms, namely augmented label-smoothed contrastive loss $\mathcal{L}_{\text{sc}}$ (Section \ref{sec:soft-contrast}), category aware ViT loss $\mathcal{L}_{\text{ct}}$ (Section \ref{sec:attn-prune}), and neighborhood-regularized mask language modeling loss $\mathcal{L}_{\text{r-mlm}}$ (Section \ref{sec:knn}) correspond to each of the three prementioned modules respectively. 

\subsection{Augmented Label-Smoothed Contrast for Multi-modality Loose Alignment (S1)}
\label{sec:soft-contrast}
Contrastive objectives have been proven effective in multimodal pre-training \citep{radford2021learning} by minimizing the representation distance between different modalities of the same data point while keeping those of different samples away~\cite{yu2022coco}. However, for attribute value extraction, the image and textual descriptions are typically \textit{loosely aligned} from two perspectives: (1) \textit{Intra-sample weak alignment}: The text description may not necessarily form a coherent and complete sentence, but a set of semantic fragments describing multiple facets. Thus, grounding the language to corresponding visual regions is difficult. (2) \textit{Potential inter-samples alignment}: Due to the commonality of samples, the textual description of one sample may also correspond to the image of another. Thus, traditional binary matching and contrastive objectives become suboptimal for these loosely-aligned texts and images.

To handle the looseness of images and texts, we augment the contrast to include sample comparison outside the batch with two queues storing the most recent $M$ ($M$ $\gg$ batch size $B$ ) visual and textual representations, inspired by the momentum contrast in MoCo \citep{he2020momentum} and ALBEF \citep{li2021align}. 
For the \textit{intra-sample weak alignment} of each given sample $\mathcal{X}_n$, instead of using the one-hot pairing label $\mathbf{p}^{i2t}_n$, we smooth the pairing target with the pseudo-similarity $\mathbf{q}^{i2t}_n$, 
\begin{equation}
\small
\setlength{\abovedisplayskip}{3pt}
\setlength{\belowdisplayskip}{3pt}
\label{equ:soft-target}
\mathbf{\widetilde{p}}^{i2t}_n = (1-\alpha)\mathbf{p}^{i2t}_n + \alpha\mathbf{q}^{i2t}_n,
\end{equation} 
where $\alpha$ is a hyper-parameter and $\mathbf{q}^{i2t}_n$ is calculated by softmax over the representation multiplication of the \verb|[CLS]| tokens,  $\bm v_n^{'\text{cls}}$ and $\bm t_n^{'\text{cls}}$, from momentum unimodal encoders $\mathcal{F'}_v$ and $\mathcal{F'}_t$,
\begin{equation}
\small
\setlength{\abovedisplayskip}{3pt}
\setlength{\belowdisplayskip}{3pt}
\mathbf{q}^{i2t}_n = \sigma\left({\mathcal{F'}_v\left(\mathcal{I}_n\right)}^{\top} \mathcal{F'}_t\left(\mathcal{T}_n\right)\right) = \sigma\left({\bm v_n^{'\text{cls}}} ^{\top} \bm t_n^{'\text{cls}} \right).
\end{equation}
For \textit{potential inter-samples pairing relations}, the visual representation $\bm v_n^{'\text{cls}}$ is compared with all textual representations $\boldsymbol{T}'$ in the queue to augment contrastive loss. Formally, the predicted image-to-text matching probability of $\mathcal{X}_n$ is
\begin{equation}
\small
\setlength{\abovedisplayskip}{3pt}
\setlength{\belowdisplayskip}{3pt}
\mathbf{d}^{i2t}_n = \frac{\exp \left({\bm v_n^{'\text{cls}}}^{\top} \boldsymbol{T}'_{m}/ \tau\right)}{\sum_{m=1}^M \exp \left({\bm v_n^{'\text{cls}}}^{\top} \boldsymbol{T}'_{m}/ \tau\right)}.
\label{eq:tem}
\end{equation}
With the smoothed targets from Equation (\ref{equ:soft-target}), the \textit{image-to-text} contrastive loss $L_{i2t}$ is calculated as the cross-entropy between the smoothed targets $\mathbf{\widetilde{p}}^{i2t}_n$ and contrast-augmented predictions $\mathbf{d}^{i2t}_n$,
\begin{equation}
\small
\setlength{\abovedisplayskip}{3pt}
\setlength{\belowdisplayskip}{3pt}
L_{i2t}=-\frac{1}{N}\left(\sum_{n=1}^N \mathbf{\widetilde{p}}_{n}^{i2t} \cdot \log \left(\mathbf{d}_{n}^{i2t}\right)\right),
\end{equation}
and vise versa for the \textit{text-to-image} contrastive loss $L_{t2i}$. Finally, the augmented label-smoothed contrastive loss $L_{\text{sc}}$ is the average of these two terms, 
\begin{equation}
\small
\setlength{\abovedisplayskip}{3pt}
\setlength{\belowdisplayskip}{3pt}
L_{\text{sc}} = \left(L_{i2t} + L_{t2i}\right) / 2.
\end{equation}

\subsection{Visual Attention Pruning (S2)}
\label{sec:attn-prune}
Images usually contain not only the visual foreground of the concerned category but also rich background contexts. 
Although previous studies indicate context can serve as an effective cue for visual understanding \cite{doersch2015unsupervised,zhang2020putting, DBLP:conf/iclr/XiaoEIM21}, it has been found that the output of ViT is often based on supportive signals in the background rather than the actual object \citep{chefer2022optimizing}. Especially in a fine-grained task such as attribute value extraction, the associated backgrounds could distract the visual model and harm the prediction precision. For example, when predicting the color of \textit{birthday balloons}, commonly co-occurring contexts such as \textit{flowers} could mislead the model and result in wrongly predicted values. 

To encourage the ViT encoder $\mathcal{F}$ focus on task-relevant foregrounds given the input image $\mathcal{I}_n$, we add a category-aware attention pruning schema, supervised with category classification, 
\begin{equation}
\small
\setlength{\abovedisplayskip}{3pt}
\setlength{\belowdisplayskip}{3pt}
L_{\text{ct}}=-\frac{1}{N}\left(\sum_{n=1}^N c_n \cdot \log \left(\mathcal{F}(\mathcal{I}_n)\right)\right).
\end{equation}
In real-world information extraction tasks, `category' denote classification schemas for organizing and structuring diverse data, exemplified by the broad range of product types in e-commerce, such as electronics, clothing, or books. These categories not only display vast diversity but also have distinct data distributions and properties, adding layers of complexity to the information extraction scenarios.

The learned attention mask $\bm{M}$ in ViT can gradually resemble the object boundary of the interested category and distinguishes the most important task-related regions from backgrounds by assigning different attention weights to the image patches~\citep{selvaraju2017grad}. The learned $\bm{M}$ is then applied on the visual representation sequences $\bm{v}_n$ of the whole image,
\begin{equation}
\small
\setlength{\abovedisplayskip}{3pt}
\setlength{\belowdisplayskip}{3pt}
\bm{v}^{pt}_n = \bm{v}_n \odot \sigma(\bm M),
\end{equation}
to screen out noisy background and task-irrelevant patches before concatenating with the textual representation $\bm{t}_n$ for further cross-modal grounding.

\subsection{Two-level Neighborhood-regularized Sample Weight Adjustment (S3)}
\label{sec:knn}
Weak labels from established models can be noisy and biased toward the textual input. Directly training the models with these labels leads to a learning gap across modalities. Prior work on self-training shows that embedding similarity can help to mitigate the label errors issue \citep{xu2023neighborhood, lang2022training}. Inspired by this line of work, we design a two-level neighborhood-regularized sample weight adjustment. In each iteration, sample weight $s\left(\mathcal{X}_n\right)$ is updated based on its label reliability, which is then applied to the training objective of attribute value generation in the next iteration,
\begin{equation}
\setlength{\abovedisplayskip}{3pt}
\setlength{\belowdisplayskip}{3pt}
\small
\mathcal{L}_{\text{r-mlm}} = -\frac{1}{N}\left(\sum_{n=1}^N s\left(\mathcal{X}_n\right) \cdot g \left(y_n, \hat{y}_n\right)
\right),
\end{equation}
where $g$ measures the element-wise cross entropy between the training label $y_n$ and the prediction $\hat{y}_n$.
As illustrated by the right example in Figure \ref{fig:framework}\footnote{See Appendix \ref{sec:demo} for additional demo examples.}, where green arrows point to samples with the same training label as $y_n$, and red arrows point to either visual or prediction neighbors, a higher consistency between the two sets indicates a higher reliability of $y_n$, formally explained as below: 

\noindent\textbf{(1) Visual Neighbor Regularization.}
The first level of regularization is based on the consistency between the sample set with the same training label $y_n$ and visual feature neighbors of $\mathcal{X}_n$. 
For each sample $\mathcal{X}_n$ with visual representation $\bm{v}_n$, we adopt the $K$-nearest neighbors (KNN) algorithm to find its neighbor samples in the visual feature space: 
\begin{equation}
\small
\setlength{\abovedisplayskip}{3pt}
\setlength{\belowdisplayskip}{3pt}
\mathcal{N}_n=\left\{\mathcal{X}_n  \cup \mathcal{X}_k \in \operatorname{KNN}\left(\bm{v}_n, \mathcal{D}, K\right)\right\},
\end{equation}
where $\operatorname{KNN}\left(\bm{v}_n, \mathcal{D}, K\right)$ demotes $K$ samples in $\mathcal{D}$ with visual representation nearest to $\bm{v}_n$. 
Simultaneously, we obtain the set of samples in $\mathcal{D}$ with the same training label $y_j$ as that of the sample $\mathcal{X}_n$,
\begin{equation}
\small
\setlength{\abovedisplayskip}{3pt}
\setlength{\belowdisplayskip}{3pt}
\mathcal{Y}_n=\left\{\mathcal{X}_n \cup \mathcal{X}_j \in \mathcal{D}_{y_j=y_n}\right\}.
\label{equ:consensus}
\end{equation}
The reliability of sample $\mathcal{X}_n$ based on the visual neighborhood regularization is 
\begin{equation}
\small
\setlength{\abovedisplayskip}{3pt}
\setlength{\belowdisplayskip}{3pt}
s_{v}(\mathcal{X}_n) = \left|\mathcal{N}_n \cap \mathcal{Y}_n\right| / {K}.
\end{equation}

\noindent\textbf{(2) Prediction Neighbor Regularization.}
The second level of regularization is based on the consistency between the sample set with the same training label and the prediction neighbors from the previous iteration, which represents the learned multimodal representation. Prediction regularization is further added after $E$ epochs when the model can give relatively confident predictions, ensuring the predicted values are qualified for correcting potential noise.
Formally, we obtain the set of samples in $\mathcal{D}$ whose predicted attribute value $p_j$ from the last iteration is the same as that of the sample $\mathcal{X}_n$, 
\begin{equation}
\small
\setlength{\abovedisplayskip}{3pt}
\setlength{\belowdisplayskip}{3pt}
\hat{\mathcal{Y}}_n=\left\{\mathcal{X}_n \cup \mathcal{X}_j \in \mathcal{D}_{\hat{y}_j=\hat{y}_n}\right\}.
\end{equation}
With the truth-value consensus set $\mathcal{Y}_n$ from Equation~(\ref{equ:consensus}), the reliability based on previous prediction neighbor regularization of the sample $\mathcal{X}_n$ is 
\begin{equation}
\small
\setlength{\abovedisplayskip}{3pt}
\setlength{\belowdisplayskip}{3pt}
s_{p}\left(\mathcal{X}_n\right) = \left|\hat{\mathcal{Y}}_n \cap \mathcal{Y}_n\right|/\left|\hat{\mathcal{Y}}_n \cup \mathcal{Y}_n\right|.
\end{equation}
Overall, $s(\mathcal{X}_n)$ is initially regularized with visual neighbors and jointly with prediction neighbors after $E$ epochs when the model predicts credibly,
\begin{equation}
\setlength{\abovedisplayskip}{3pt}
\setlength{\belowdisplayskip}{3pt}
\small
s\left(\mathcal{X}_n\right) = \left\{
\begin{aligned}
& s_{v}\left(\mathcal{X}_n\right)  & e < E, \\
& \text{AVG}\left(s_{v}\left(\mathcal{X}_n\right), s_{p}\left(\mathcal{X}_n\right)\right) & e \geq E.
\end{aligned}
\right.
\end{equation}

\begin{table}
\centering
\resizebox{\columnwidth}{!}{%
\begin{tabular}{cccccc}
\toprule
\textbf{Attr} & \textbf{\# PT} & \textbf{Value Type}  & \textbf{\# Valid} & \textbf{\# Train \& Val} & \textbf{\# Test}   \\
\midrule
Item Form & 14 & Single & 142 & 42,911 & 4,165 \\
Color & 255 & Multiple  & 24 & 106,176 & 3,777 \\
Pattern & 31  & Single & 30 & 119,622 & 2,093 \\
\bottomrule
\end{tabular}
}
\caption{Statistics of the attribute extraction datasets.}
\label{tab:dataset}
\end{table}

\begin{table*}
\centering
\begin{adjustbox}{max width=\textwidth}
\begin{tabular}{cccccccccccccc}
\toprule
\multirow{2.5}{*}{Type} & \multirow{2.5}{*}{Method} &\multicolumn{3}{c}{\bf Dataset: Item Form} & & \multicolumn{3}{c}{\bf Dataset: Color}& & \multicolumn{3}{c}{\bf Dataset: Pattern}\\
\cmidrule(lr){3-5} \cmidrule(lr){7-9} \cmidrule(lr){11-13} 
& & {Precision} & {Recall} & {$\text{F}_1$} & { } & {Precision} & {Recall} & {$\text{F}_1$} & { } & {Precision} & {Recall} & {$\text{F}_1$} \\
\midrule
\multirow{3}{*}{Unimodal}
& $\text{OpenTag}_\text{seq}$ & 91.37 & 44.97 & 60.27 & & 83.94 & 24.73 & 38.20 & & 79.65 & 19.83 & 31.75 \\
& $\text{OpenTag}_\text{cls}$ & 89.40 & 51.67 & 65.49 & & 81.13 & 28.61 & 42.30 & & 78.10 & 24.41 & 37.19 \\
& TEA & 82.71 & 60.98 & 70.20 & & 67.58 & 47.80 & 55.99 & & 60.87 & 37.40 & 46.33 \\
\midrule
\multirow{5}{*}{Multimodal}
& ViLBERT & 75.97 & 65.67 & 70.45 & & 60.22 & 51.12 & 55.30 & & 60.10 & 40.52 & 48.40 \\
& LXMERT & 75.79 & 68.72 & 72.08 & & 60.20 & 54.26 & 57.08 & & 60.33 & 42.28 & 49.72 \\
& UNITER & 76.75 & 69.10 & 72.72 & & 61.30 & 54.69 & 57.81 & & 62.45 & 43.38 & 51.20 \\
& BLIP & 78.21 & 69.25 & 73.46 & & 62.70 & 58.23 & 60.38 & & 58.74 & 44.01 & 50.32 \\
& PAM & 78.83 & 74.35 & \underline{76.52} & & 63.34 & 60.43 & \underline{61.85} & & 61.80 & 44.29 & \underline{51.60} \\
\midrule
\multirow{4}{*}{Ours} 
& \ours w/o S1 & 80.03 & 72.49 & 76.07 & & 71.00 & 58.41 & 64.09 & & 60.03 & 45.59 & 51.82 \\
& \ours w/o S2 & 80.48 & 75.32 & 77.81 & & 73.77 & 59.37 & 65.79 & & 59.01 & 46.74 & 52.16 \\
& \ours w/o S3 & 80.87 & 72.71 & 76.57 & & 74.29 & 59.04 & 65.79 & & 59.92 & 44.92 & 51.35 \\
& \cellcolor{gray!20} \ours & \cellcolor{gray!20}82.46 & \cellcolor{gray!20}75.40 & \cellcolor{gray!20}\textbf{78.77} & \cellcolor{gray!20}  & \cellcolor{gray!20}77.44 & \cellcolor{gray!20}60.19 & \cellcolor{gray!20}\textbf{67.73} & \cellcolor{gray!20} & \cellcolor{gray!20}62.10 & \cellcolor{gray!20}46.84 & \cellcolor{gray!20}\textbf{53.40} \\
\bottomrule
\end{tabular}
\end{adjustbox}
\caption{Performance comparison with different baselines (\%). The performance gains over the baselines have passed the t-test with a p-value$<$0.05. The best performance is in bold, and the second runner baseline is underlined.}
\label{tab:overall-performance}
\end{table*}
\section{Experimental Setup}
\label{sec:exp-setup}
\subsection{Dataset and Implementation Details}
\label{sec:dataset}
We build three multimodal attribute value extraction datasets by collecting profiles (title, bullets, and descriptions) and images from the public \verb|amazon.com| web pages,
where each dataset corresponds to one attribute $\mathcal{R}$. The dataset information is summarized in Table \ref{tab:dataset}, where \textbf{Attr} is the attribute $\mathcal{R}$, \textbf{\# PT} represents the number of unique categories (i.e., product types), \textbf{Value Type} indicates whether $y_n$ contain single or multiple values, and \textbf{\# Valid} represents the number of valid values. To better reflect real-world scenarios, we use the attribute-value pairs from the product information section on web pages as 
weak training labels instead of highly processed data. We follow the same filtering strategy from prior text established work~\citep{zalmout2022prototype} to denoise training data. 
For the testing, we manually annotate gold labels on the benchmark dataset to ensure preciseness. Besides, the label sources are marked down, indicating whether the attribute value is present or absent in the text, to facilitate fine-grained source-aware evaluation. The human-annotated benchmark datasets will be released to encourage the future development of modality-balanced multimodal extraction models. %
See Appendix \ref{sec:implement} for the implementation and computation details of \ours.

\subsection{Evaluation Protocol}
We use Precision, Recall, and F1 score based on synonym normalized exact string matching. For single value type, an extracted value $\hat{y}_n$ is considered correct when it exactly matches the gold value string $y_n$. For multiple value type where the gold values for the query attribute $\mathcal{R}$ can contain multiple answers $y_n \in \left\{y_n^1, \ldots, y_n^m\right\}$, the extraction is considered correct when all the gold values are matched in the prediction. Macro-aggregation is performed across attribute values to avoid the influence of class imbalance. All reported results are the average of three runs under the best settings. 

\subsection{Baselines}
We compare our proposed model with a series of baselines, spanning unimodal-based methods and multimodal-based ones. For unimodal baselines, OpenTag~\citep{zheng2018opentag} is considered a strong text-based model for attribute extraction. $\text{OpenTag}_\text{seq}$ formulates the task as sequence tagging and uses the BiLSTM-CRF architecture with self-attention. $\text{OpenTag}_\text{cls}$ replaces the BiLSTM encoder with a transformer encoder and tackles the task as classification. TEA is another text-only unimodal generative model with the same architecture as \ours but without the image patching, which is included to demonstrate the influence of the generation setting. For multimodal baselines, we consider discriminative encoder models, including ViLBERT~\citep{lu2019vilbert}, LXMERT~\citep{tan2019lxmert} with dual encoders, and UNITER~\citep{chen2020uniter} with a joint encoder. We also add generative encoder-decoder models for comparisons. BLIP~\citep{DBLP:conf/icml/0001LXH22} adopts dual encoders and an image-grounded text decoder. PAM~\citep{lin2021pam} uses a shared encoder and decoder separated by a prefix causal mask.

\section{Experimental Results}
\label{sec:exp-result}
\begin{table}
\centering
\resizebox{\columnwidth}{!}{%
\begin{tabular}{cccccc}
\toprule
Method & Gold Value Source &{Precision} & {Recall} & {$\text{F}_1$} \\
\midrule
\multirow{3}{*}{$\text{OpenTag}_\text{cls}$}
& Text \cmark & 89.78 & 52.13 & 65.96  \\
& Text \xmark \quad Image \cmark & 78.95 & 31.25 & 44.78 \\
\cdashline{2-5}[0.8pt/2pt]
& \cellcolor{red!10} \bf GAP $\downarrow$ & \cellcolor{red!10} 10.83 &\cellcolor{red!10}  20.88 & \cellcolor{red!10} 21.18 \\
\midrule
\multirow{3}{*}{PAM}
& Text \cmark  & 79.16 & 74.53 & 76.78 \\
& Text \xmark \quad Image \cmark & 66.67 & 58.33 & 62.22 \\
\cdashline{2-5}[0.8pt/2pt]
& \cellcolor{red!10} \bf GAP $\downarrow$ & \cellcolor{red!10} 12.50 & \cellcolor{red!10} 16.20 & \cellcolor{red!10} \underline{14.56} \\
\midrule
\multirow{3}{*}{\ours}
& Text \cmark  & 82.64 & 75.71 & 79.02 \\
& Text \xmark \quad Image \cmark & 75.00 & 62.50 & 68.18 \\
\cdashline{2-5}[0.8pt/2pt]
& \cellcolor{green!10} \bf GAP $\downarrow$ & \cellcolor{green!10} 7.64 & \cellcolor{green!10} 13.21 & \cellcolor{green!10} \textbf{10.84} \\
\bottomrule
\end{tabular}
}
\caption{Fine-grained source-aware evaluation of different methods. The \textit{gold value source} indicates whether the gold value is contained in the text, or is not contained in the text and must be inferred from the image.}
\label{tab:source-aware}
\end{table}
\subsection{Overall Comparison}
Table \ref{tab:overall-performance} shows the performance comparison of different types of extraction methods. It is shown that \ours achieves the best $\text{F}_1$ performance, especially compared to unimodal baselines, demonstrating the advantages of patching visual modality to this text-established task. Comparing the unimodal methods with multimodal ones, textual-only models achieve impressive results on precision while greatly suffering from low recall, which indicates potential information loss when the gold value is not contained in the input text. With the generative setting, TEA sort of mitigates the information loss and improves recall over OpenTag under the tagging and classification settings. Besides, adding visual information can further improve recall, especially for the multi-value attribute \textit{Color}, where multimodal models can even double that of text-only ones. However, the lower precision performance of the multimodal models implies the challenges beneath cross-modality integration. With the three proposed bias-reduction schemes, \ours improves on all three metrics over multimodal baselines and balances precision and recall to a great extent compared with unimodal models. Besides the full \ours, we also include three variants that remove one proposed schema at a time. It shows that the visual attention pruning module mainly helps with precision while the other two benefit both precision and recall, leading to the best $\text{F}_1$ performance when all three schemes are equipped. We include several case studies in Section \ref{sec:case-study} for qualitative observation.

\noindent\textbf{Source-Aware Evaluation.}
\label{sec:source-aware-eva}
To investigate how the modality learning bias is addressed, we conduct fine-grained source-aware evaluation similarly to Section \ref{sec:motivation}, as shown in Table \ref{tab:source-aware}\footnote{We demonstrate results on the Item Form dataset due to limited space. For more results, please refer to Appendix \ref{sec:source-aware-more}.}. 
The performance gap between when the gold value is present or absent in the text is significantly reduced by \ours when compared to both unimodal and multimodal representative methods, which suggests a more balanced and generalized capacity of \ours to learn from different modalities.
When the gold value is absent in the text, our method outperforms $\text{OpenTag}_\text{cls}$ by more than twice as much on recall, and also outperforms on precision under various scenarios compared to the multimodal PAM.

\subsection{Ablation Studies}
\label{sec:abla}
\noindent\textbf{Augmented Label-Smoothed Contrast.}
\label{sec:abla-sc}
\begin{table}
\centering
\resizebox{\columnwidth}{!}{%
\begin{tabular}{ccccccccc}
\toprule
\multirow{2.5}{*}{Method}  & \multicolumn{3}{c}{\bf Single Value Dataset} & & \multicolumn{3}{c}{\bf Multiple Value Dataset}\\
\cmidrule(lr){2-4} \cmidrule(lr){6-8} 
& {P} & {R} & {$\text{F}_1$} & { } & {P} & {R} & {$\text{F}_1$}  \\
\midrule
w/o $L_\text{sc}$ & 80.03 & 72.49 & 76.07 && 71.00 & 58.41 & 64.09 \\
w/o Smooth  & 81.42 & 74.41 & 77.76 && 75.06 & 59.99 & 66.68 \\
\rowcolor{gray!20} \ours & 82.46 & 75.40 & 78.77 && 77.44 & 60.19 & 67.73 \\
\bottomrule
\end{tabular}
}
\caption{Ablation study on the augmented label-smoothed contrast for cross-modality alignment (\%).}
\label{tab:abla_sc}
\end{table}
\begin{figure}
    \centering
    \includegraphics[width=\linewidth]{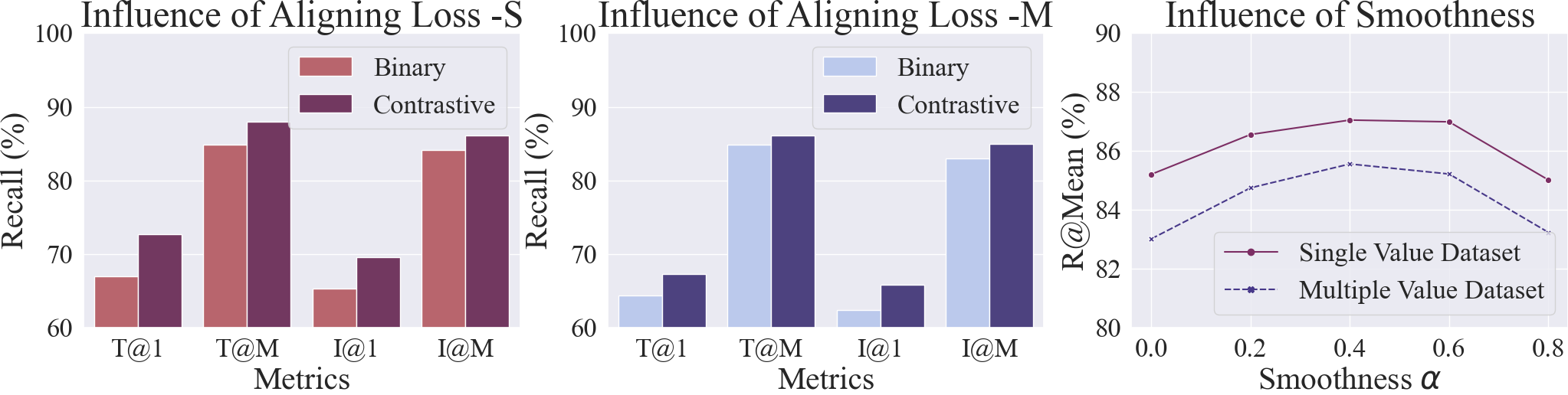}
    \caption{The influence study of alignment objectives, i.e., binary matching v.s. contrastive loss, and the influence of softness $\alpha$ via the task of image-to-text and text-to-image retrieval. The metric T/I@1 is the recall of text/image retrieval at rank 1, T/I@M means the rank average, and R@Mean further averages T@M and I@M. }
    \label{fig:retrieval}
\end{figure}
We look into the impact of label-smoothed contrast on both single- and multiple-value type datasets \footnote{For ablation analysis, we select Item Form as the representative for single-value and Color for multiple-value type dataset. More ablation results can be referred in Appendix \ref{sec:ablation-more}.}. 
Table \ref{tab:abla_sc} shows that removing the contrastive objective leads to a drop in both precision and recall. For the multiple-value dataset, adding the contrastive objective significantly benefits precision, suggesting it encourages cross-modal validation when there are multiple valid answers in the visual input. With label smoothing, the recall can be further improved. This indicates that the augmented and smoothed contrast can effectively leverage the cross-modality alignment inter-samples, hence improving the coverage rate when making predictions.

In addition, we conduct cross-modality retrieval to study the efficacy of aligning objectives, i.e., binary matching and contrastive loss, for cross-modality alignment and the influence of the softness $\alpha$, as shown in Figure \ref{fig:retrieval}. Across different datasets and metrics, the contrastive loss consistently outperforms the binary matching loss. This consolidates our choice of contrasting objectives and highlights the potential benefits of label-smoothing and contrast augmentation, given that both are neglected in a binary matching objective. Retrieval performance under different smoothness values shows a trend of first rising and then falling. We simply take 0.4 for $\alpha$ in our experiments. 

\noindent\textbf{Category Aware Attention Pruning.}
\begin{table}
\centering
\resizebox{\columnwidth}{!}{%
\begin{tabular}{ccccccccc}
\toprule
\multirow{2.5}{*}{Method} & \multicolumn{3}{c}{\bf Single Value Dataset}& & \multicolumn{3}{c}{\bf Multiple Value Dataset}\\
\cmidrule(lr){2-4} \cmidrule(lr){6-8} 
& {P} & {R} & {$\text{F}_1$} & { } & {P} & {R} & {$\text{F}_1$} \\
\midrule
w/o $L_\text{ct}$  & 80.48 & 75.32 & 77.81 && 73.77 & 59.37 & 65.79 \\
w/o Attn Prun & 80.61 & 75.49 & 77.97 && 74.60 & 59.42 &  66.15 \\
\rowcolor{gray!20} \ours & 82.46 & 75.40 & 78.77 && 77.44 & 60.19 & 67.73 \\
\bottomrule
\end{tabular}
}
\caption{Ablation study on the category supervised visual attention pruning (\%).}
\label{tab:abla-attn_prun}
\end{table}
\begin{figure}
    \centering
    \includegraphics[width=\linewidth]{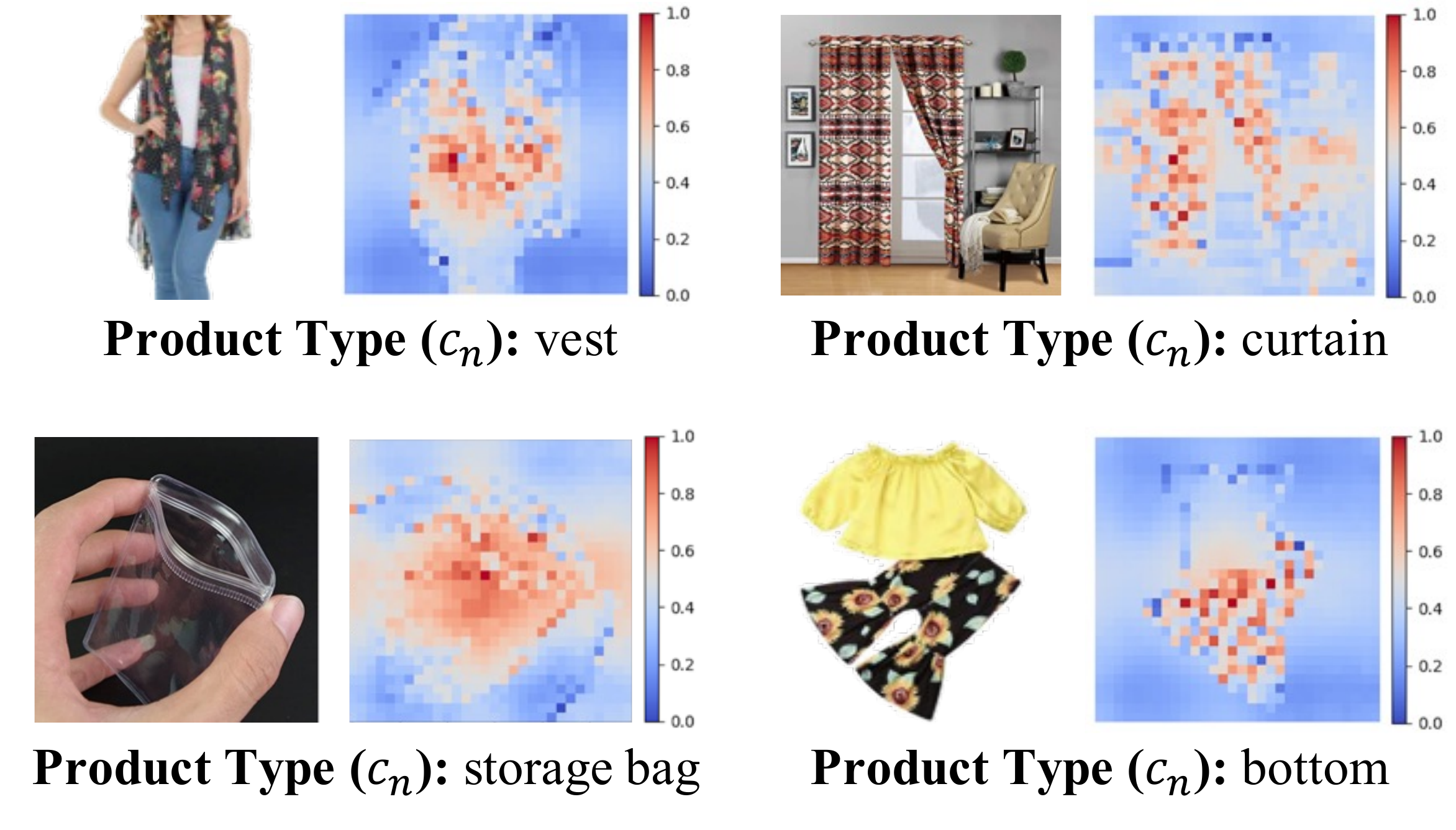}
    \caption{Visualization of learned attention mask with category (e.g., product type) aware ViT classification.}
    \label{fig:attn-prune}
\end{figure}
We study the influence of the category aware attention pruning, as shown in Table \ref{tab:abla-attn_prun}. The results imply that adding the category classification helps to improve precision performance without harming recall, and the learned attention mask can effectively highlight the foreground regions of the queried sample. Figure \ref{fig:attn-prune} presents several visualizations of the learned attention mask. 

\noindent\textbf{Neighborhood Regularization.}
\begin{table}
\centering
\resizebox{\columnwidth}{!}{%
\begin{tabular}{ccccccccc}
\toprule
\multirow{2.5}{*}{Method} & \multicolumn{3}{c}{\bf Single Value Dataset}& & \multicolumn{3}{c}{\bf Multiple Value Dataset}\\
\cmidrule(lr){2-4} \cmidrule(lr){6-8} 
& {P} & {R} & {$\text{F}_1$} & { } & {P} & {R} & {$\text{F}_1$} \\
\midrule
w/o NR & 80.87 & 72.71 & 76.57 && 74.29 & 59.04 & 65.79 \\
w/o Vis-NR & 81.87 & 73.54 & 77.48 && 77.07 & 59.99 & 67.47 \\
w/o Pred-NR & 81.81 & 73.18 & 77.25 && 76.71 & 59.44 & 66.98 \\
\rowcolor{gray!20} \ours & 82.46 & 75.40 & 78.77 && 77.44 & 60.19 & 67.73 \\
\bottomrule
\end{tabular}
}
\caption{Ablation study on the two-level neighborhood-regularized sample weight adjustment (\%).}
\label{tab:abla-neigh_reg}
\end{table}
We consider the influence of the two-level neighborhood regularization by removing the visual neighborhood regularization (Vis-NR), prediction neighborhood regularization (Pred-NR), or both (NR) from the full model.
Results in Table \ref{tab:abla-neigh_reg} show all the metrics decrease when both regularizations are removed, indicating the validity of the proposed neighborhood regularized sample weight adjustment in mitigating the influence of hard, noisy samples. Besides, since the second-level prediction-based neighbor regularization is independent of the multimodal extraction framework, it can be incorporated flexibly into other frameworks as well for future usage. 

\noindent\textbf{Classification vs. Generation}
\label{sec:cla-gen}
\begin{table}
\centering
\resizebox{\columnwidth}{!}{%
\begin{tabular}{ccccccccccccc}
\toprule
\multirow{2.5}{*}{\bf Setting} &\multicolumn{3}{c}{$\mathcal{D}$: \bf Item Form} & & \multicolumn{3}{c}{$\mathcal{D}$: \bf Color}& & \multicolumn{3}{c}{$\mathcal{D}$: \bf Pattern}\\
\cmidrule(lr){2-4} \cmidrule(lr){6-8} \cmidrule(lr){10-12} 
& {P} & {R} & {$\text{F}_1$} & { } & {P} & {R} & {$\text{F}_1$} & { } & {P} & {R} & {$\text{F}_1$} \\
\midrule
Classification & 79.93 & 70.47 & 74.90 & & 72.21 & 50.18 & 59.21 & & 59.08 & 42.16 & 49.21 \\
\rowcolor{orange!20} Generation & 82.46 & 75.40 & 78.77 & & 77.44 & 60.19 & 67.73 & & 62.10 & 46.84 & 53.40 \\
\bottomrule
\end{tabular}
}
\caption{Attribute extraction performance comparison between the settings of classification and generation.}
\label{tab:cla-gen}
\end{table}
To determine which architecture is better for multimodal attribute value extraction, we compare the generation and classification settings for the module of the attribute information extractor. The results are demonstrated in Table \ref{tab:cla-gen}. It is shown that the setting of generation achieves significant advantages over classification. Especially on the recall performance for multi-value type attribute Color, where the gold value can be multiple, the improvement of recall can be up to 20\% relatively. This indicates that the generation setting can extract more complete results from the multimodal input, leading to a higher coverage rate. Therefore, we choose the generation setting in the attribute value extraction module in the final architecture design of \ours.

\subsection{Case Study}
\label{sec:case-study}
\begin{figure}[htbp]
    \centering
    \includegraphics[width=\linewidth]{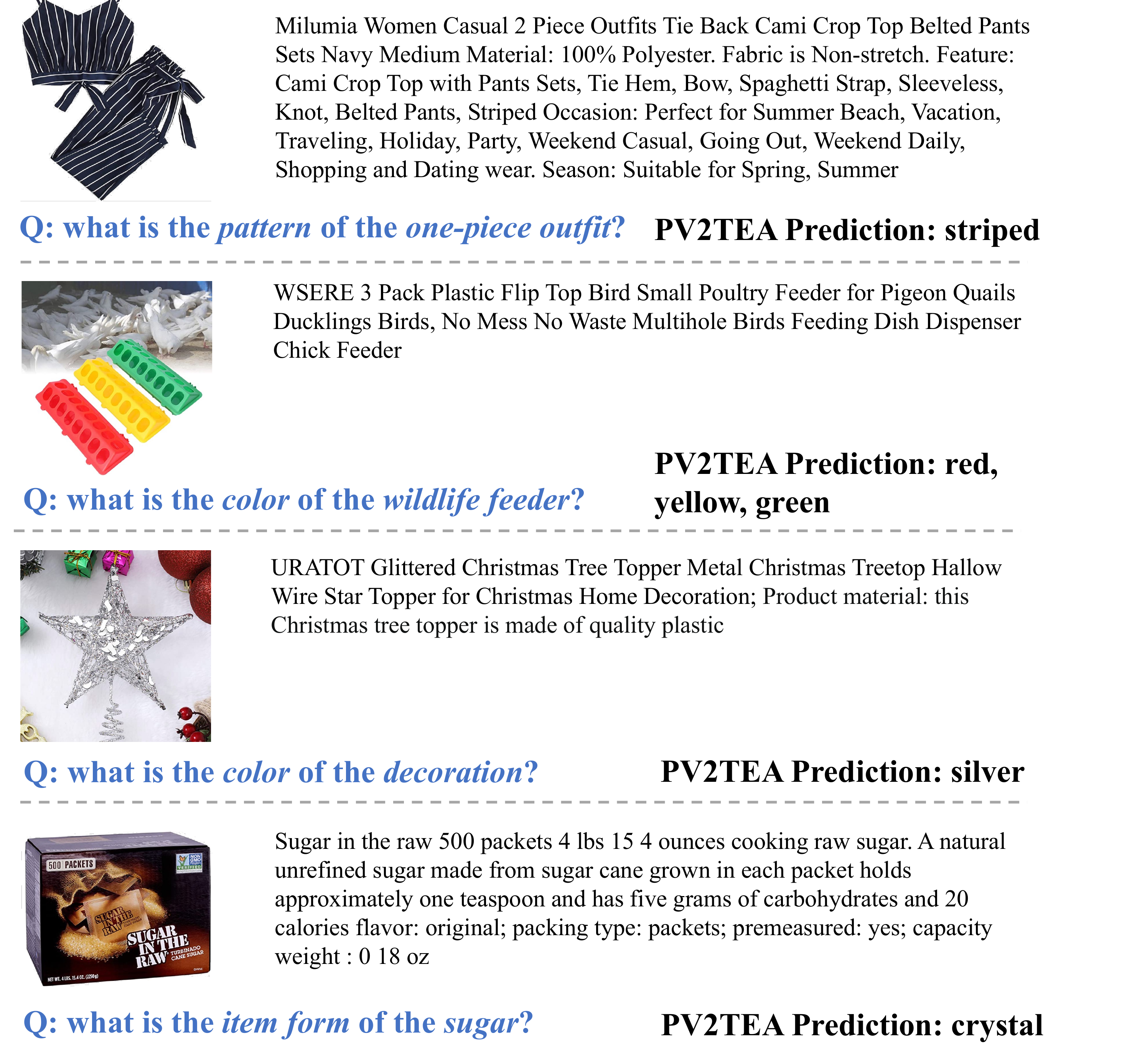}
    \caption{Qualitatively case studies.}
    \label{fig: case-study}
\end{figure}
To qualitatively observe the extraction performance, we attach several case studies in Figure \ref{fig: case-study}. It shows that even when the attribute value is not contained in the text, \ours can still perform the extraction reliably from images. In multiple value datasets such as Color, \ours can effectively differentiate related regions and extract multiple values with comprehensive coverage. 

\section{Related Work}
\label{sec:related}
\noindent\textbf{Attribute Information Extraction.}
Attribute extraction has been extensively studied in the literature primarily based on textual input. OpenTag~\citep{zheng2018opentag} formalizes it as a sequence tagging task and proposes a combined model leveraging bi-LSTM-CRF, and attention to perform end-to-end tagging. \citet{xu2019scaling} scales the sequence-tagging-based model with a global set of BIO tags. AVEQA~\citep{wang2020learning} develops a question-answering model by treating each attribute as a question and extracting the best answer span from the text. TXtract~\citep{DBLP:conf/acl/KaramanolakisMD20} uses a hierarchical taxonomy of categories and improves value extraction through multi-task learning. AdaTag~\citep{DBLP:conf/acl/YanZLGRD20} exploits an adaptive CRF-based decoder to handle multi-attribute value extractions. 
Additionally, there have been a few attempts at multimodal attribute value extraction. M-JAVE~\citep{DBLP:conf/emnlp/ZhuWLWHZ20} introduces a gated attention layer to combine information from the image and text. PAM~\citep{lin2021pam} proposes a transformer-based sequence-to-sequence generation model for multimodal attribute value extraction. Although the latter two use both visual and textual input, they fail to account for possible modality bias and are fully supervised. 

\noindent\textbf{Multi-modality Alignment and Fusion.}
The goal of multimodal learning is to process and relate information from diverse modalities. 
CLIP~\citep{radford2021learning} makes a gigantic leap forward in bridging embedding spaces of image and text with contrastive language-image pretraining. ALBEF~\citep{li2021align} applies a contrastive loss to align the image and text representation before merging with cross-modal attention, which fits loosely-aligned sample image and text. Using noisy picture alt-text data, ALIGN~\citep{jia2021scaling} jointly learns representations applicable to either vision-only or vision-language tasks. The novel Vision-Language Pre-training (VLP) framework established by BLIP~\citep{DBLP:conf/icml/0001LXH22} is flexibly applied to both vision-language understanding and generation tasks. GLIP~\citep{li2022grounded} offers a grounded language-image paradigm for learning semantically rich visual representations. FLAVA~\citep{singh2022flava} creates a foundational alignment that simultaneously addresses vision, language, and their interconnected multimodality. Flamingo~\citep{alayrac2022flamingo} equips the model with in-context few-shot learning capabilities. SimVLM~\citep{wang2021simvlm} is trained end-to-end with a single prefix language modeling and investigates large-scale weak supervision. Multi-way Transformers are introduced in BEIT-3~\citep{wang2022image} for generic modeling and modality-specific encoding.
\section{Conclusion}
\label{sec:conclu}
In this work, we propose \ours, a bias-mitigated visual modality patching-up model for multimodal information extraction. Specifically, we take attribution value extraction as an example for illustration. Results on our released source-aware benchmarks demonstrate remarkable improvements: the augmented label-smoothed contrast promotes a more accurate and complete alignment for loosely related images and texts; the visual attention pruning improves precision by masking out task-irrelevant regions; and the neighborhood-regularized sample weight adjustment reduces textual bias by lowering the influence of noisy samples. We anticipate the investigated challenges and proposed solutions will inspire future scenarios where the task is first established on the text and then expanded to multiple modalities. 
\section*{Limitations}
\label{sec:limit}
There are several limitations that can be considered for future improvements: (1) In multimodal alignment and fusion, we only consider a single image for each sample, whereas multiple images can be available. A more flexible visual encoding architecture that can digest an indefinite number of input images can improve the visual information coverage; 2) The empirical results in this work focus on three attribute extraction datasets (i.e., item form, color, and pattern) that can clearly benefit from visual perspectives, while there are also various attribute types that rely more on the textual input. Different traits of attributes may influence the preferred modalities during the modeling, which is out of scope for this work but serves as a natural extension of this study; 3) Currently there is no specific design to improve the efficiency based on the visual question answering architecture. It can be not scalable as the number of attributes increases.

There could be a dual-use regarding the attention-pruning mechanism, which can be a potential risk of this work that could arise and harm the result. The attention-pruning mechanism encourages the model to focus on the task-relevant foreground on the given image selected with category supervision, which can improve the prediction precision given the input image is visually rich and contains noisy context background. While for some types of images, such as infographics, there may be helpful text information on the images or intentionally attached by providers. These additional texts may be overlooked by the attention-pruning mechanism, resulting in potential information losses. A possible mitigation strategy is to add an OCR component along with the visual encoder to extract potential text information from given images. 

\section*{Ethics Statement}
\label{sec:ethical}

We believe this work has a broader impact outside the task and datasets in the discussion. The studied textual bias problem in our motivating analysis and the potential of training a multimodal model with weakly-supervised labels from text-established models are not restricted to a specific task. Also, it becomes common in the NLP domain that some tasks first established based on pure text input are expected to further include the consideration multimodal input. The discussion in this work can be generalized to a lot of other application scenarios. The proposed solutions for multimodal integration and modality bias mitigation are independent of model architecture, which we expect can be applied to other downstream tasks or inspire designs with similar needs.  

Regarding the human annotation involved in this work, we create three benchmark datasets that are manually labeled by human laborers to facilitate the source-aware evaluation. The annotation includes both gold attribute value as well as label sources, i.e., image or text. The profiles and images are all collected based on the publicly accessible Amazon shopping website. We depend on internal quality-assured annotators with balanced demographic and geographic characteristics, who consent and are paid adequately based in the US. The data collection protocol is approved by the ethics review board. We attach detailed human annotation instructions and usage explanations provided to the annotators in Appendix \ref{sec:annotation} for reference. 

\section*{Acknowledgements}
We would like to thank Binxuan Huang and Yan Liang for their insightful advice and thank anonymous reviewers for their feedback.
This work was partially supported by Amazon.com Services LLC, internal funds by the Computer Science Department of Emory University, and the University Research Committee of Emory University. 

\bibliography{custom}
\bibliographystyle{acl_natbib}
\balance

\clearpage
\appendix
\section{Implementation Details}
\label{sec:implement}
Our models are implemented with PyTorch~\citep{paszke2019pytorch} and Huggingface Transformer library and trained on an 8 Tesla V100 GPU node. The model is trained for 10 epochs, where the Item Form dataset takes around 12 hours, the Color dataset takes about 32 hours, and the Pattern dataset needs around 35 hours to run on a single GPU. The overall architecture of \ours consists of 361M trainable parameters, where a $\text{ViT}_\text{base}$~\citep{DBLP:conf/iclr/DosovitskiyB0WZ21} is used as the image encoder and initialized with the pre-trained model on ImageNet of 85M parameters, and the text encoder is initialized from $\text{BERT}_\text{base}$~\citep{DBLP:conf/naacl/DevlinCLT19} of 123M parameters. We use AdamW~\citep{DBLP:conf/iclr/LoshchilovH19} as the optimizer with a weight decay of 0.05. The learning rate of each parameter group is set using a cosine annealing schedule~\citep{loshchilov2016sgdr} with the initial value of 1$e$-5.
The model is trained for 10 epochs, with both training and testing batch sizes of 8. The memory queue size $M$ is set as 57600 and the temperature $\tau$ of in Equation \ref{eq:tem} is set as 0.07. We performed a grid search for the softness $\alpha$ from [0, 0.2, 0.4, 0.6, 0.8] and used the best-performed 0.4 for reporting the final results. The $K$ for two-level neighborhood regularization is set at 10. The input textual description is cropped to a maximum of 100 words. The input image is divided into 30 by 30 patches. The hidden dimension of both the visual and textual encoders is set to 768 to produce the representations of patches, tokens, or the whole image/sequence. The epoch $E$ for adding the second-level prediction neighbor regularization to reliability score $s\left(\mathcal{X}_n\right)$ is set as 2.

\section{More Source-Aware Evaluation}
\label{sec:source-aware-more}
\begin{table}[H]
\centering
\resizebox{\columnwidth}{!}{%
\begin{tabular}{cccccccccc}
\toprule
\multirow{2.5}{*}{Method} & \multirow{2.5}{*}{Gold Value Source} & \multicolumn{3}{c}{\bf $\mathcal{D}$: Color}& & \multicolumn{3}{c}{$\mathcal{D}$: \bf Pattern}\\
\cmidrule(lr){3-5} \cmidrule(lr){7-9} 
& & {P} & {R} & {$\text{F}_1$} & { } & {P} & {R} & {$\text{F}_1$}\\
\midrule
\multirow{3}{*}{$\text{OpenTag}_\text{cls}$}
& Text \cmark & 85.06 & 43.28 & 57.37 & & 85.00 & 42.96 & 57.07 \\
& Text \xmark \quad Image \cmark & 66.28 & 10.24 & 17.74 & & 66.23 & 12.02 & 20.35 \\
\cdashline{2-9}[0.8pt/2pt]
& \cellcolor{red!10} \bf GAP $\downarrow$  & \cellcolor{red!10} 18.78 & \cellcolor{red!10} 33.04 & \cellcolor{red!10} 39.63 &\cellcolor{red!10}  & \cellcolor{red!10} 18.77 & \cellcolor{red!10} 30.94 & \cellcolor{red!10} 36.72\\
\midrule
\multirow{3}{*}{PAM}
& Text \cmark & 73.20 & 71.88 & 72.53 & & 75.00 & 57.04 & 64.80 \\
& Text \xmark \quad Image \cmark & 50.30 & 45.45 & 47.75 & & 51.82 & 36.23 & 42.64\\
\cdashline{2-9}[0.8pt/2pt]
& \cellcolor{red!10} \bf GAP $\downarrow$  & \cellcolor{red!10} 22.90 & \cellcolor{red!10} 26.43 & \cellcolor{red!10} 24.78 &\cellcolor{red!10}  & \cellcolor{red!10} 23.18 & \cellcolor{red!10} 20.81 & \cellcolor{red!10} 22.16 \\
\midrule
\multirow{3}{*}{\ours} 
& Text \cmark & 81.74 & 74.25 & 77.82 & & 71.19 & 61.25 & 65.85 \\
& Text \xmark \quad Image \cmark & 71.89 & 47.19 & 56.98 & & 54.48 & 37.26 & 44.25 \\
\cdashline{2-9}[0.8pt/2pt]
& \cellcolor{green!10} \bf GAP $\downarrow$  & \cellcolor{green!10} 9.85 & \cellcolor{green!10} 27.06 & \cellcolor{green!10} \textbf{20.84} &\cellcolor{green!10}  & \cellcolor{green!10} 16.71 & \cellcolor{green!10} 23.99 & \cellcolor{green!10} \textbf{21.59} \\
\bottomrule
\end{tabular}
}
\caption{Fine-grained source-aware evaluation for the Color and Pattern datasets.}
\label{tab:source_aware_full}
\end{table}
The source-aware evaluation of the Color and Pattern datasets is shown in Table \ref{tab:source_aware_full}. We can observe that similarly to the discussions in Section \ref{sec:source-aware-eva}, compared with the baselines, the proposed \ours effectively mitigates the performance gap of $\text{F}_1$ when the gold value is not contained in the text. More specifically, we observed that compared with the unimodal method, \ours mainly reduces the recall performance gap across modalities, while compared with the multimodal method, the reduction happens mainly in precision, which all corresponds to the weaker metrics for each type of method. This indicates the stronger generalizability and more balanced learning ability of \ours.

\begin{figure*}[ht]
    \centering
    \includegraphics[width=\linewidth]{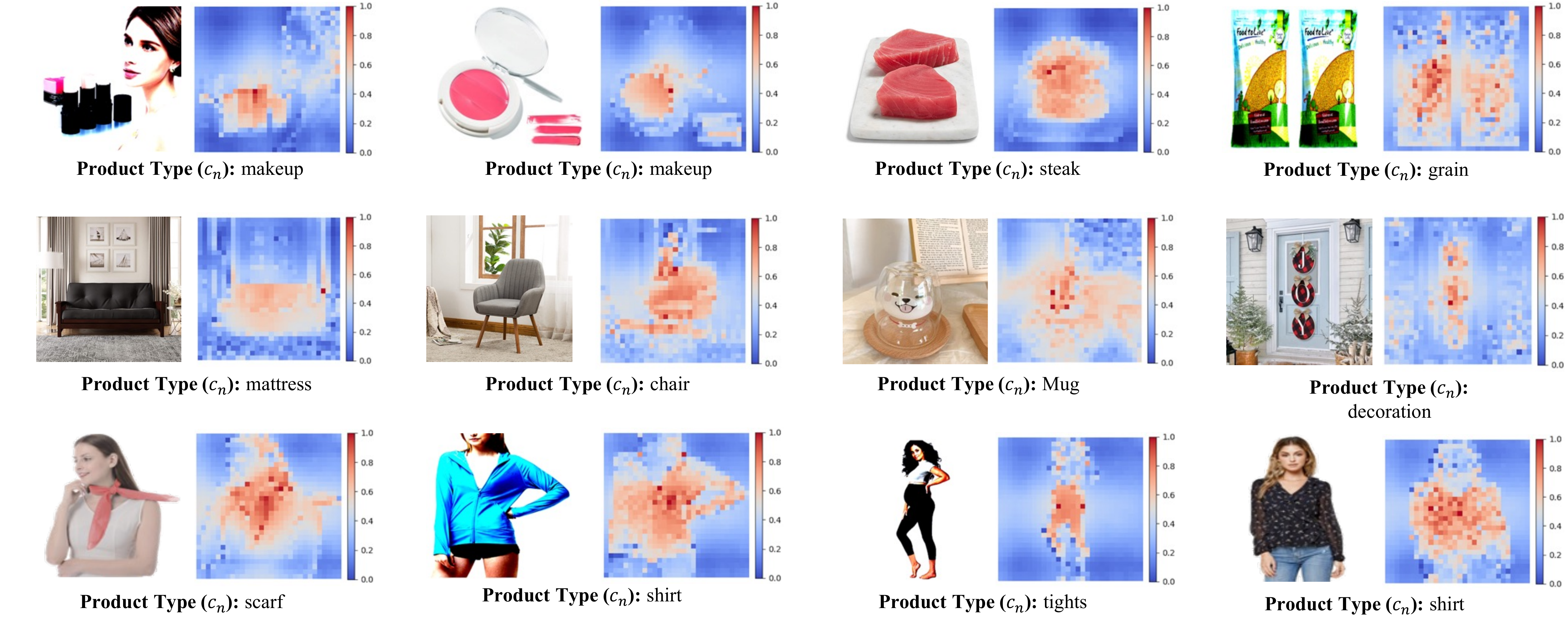}
    \caption{Visualization examples of the learned category aware attention pruning mask.}
    \label{fig:attn_vis_appendix}
\end{figure*}

\section{Ablation Studies on Pattern Dataset}
\label{sec:ablation-more}
We further include the ablation results on the single-value type dataset Pattern for each proposed mechanism in Table \ref{tab:abla_sc_full}, Table \ref{tab:abla_attn_prun_full}, and Table \ref{tab:abla_neigh_reg_full}, respectively. The observations are mostly consistent with the discussion in section \ref{sec:abla}, where all three proposed mechanisms support improvements in the overall performance of $\text{F}_1$. It is noted that the recall performance with attention-pruning drops a bit compared with that without. This may indicate potential information losses on the challenging dataset such as Pattern with only the selected foreground. We discuss this potential risk in detail in the Limitation section.

\begin{table}[H]
\centering
\small
\resizebox{\columnwidth}{!}{%
\begin{tabular}{ccccc}
\toprule
\multirow{2.5}{*}{Method} &\multicolumn{3}{c}{\bf Single Value Dataset: Pattern} \\
\cmidrule(lr){2-4} 
& {Precision} & {Recall} & {$\text{F}_1$} \\
\midrule
\ours w/o $L_\text{sc}$ & 60.03 & 45.59 & 51.82 \\
\ours w/o smooth & 61.87 & 45.72 & 52.58  \\
\rowcolor{gray!20} \ours & 62.10 & 46.84 & 53.40 \\
\bottomrule
\end{tabular}
}
\caption{Ablations on the augmented label-smoothed contrast for cross-modality alignment (\%).}
\label{tab:abla_sc_full}
\end{table}

\begin{table}[H]
\centering
\small
\resizebox{\columnwidth}{!}{%
\begin{tabular}{cccccc}
\toprule
\multirow{2.5}{*}{Method} &\multicolumn{3}{c}{\bf Single Value Dataset: Pattern} \\
\cmidrule(lr){2-4}
& {Precision} & {Recall} & {$\text{F}_1$} \\
\midrule
\ours w/o $L_\text{ct}$ \& Attn Prun & 59.01 & 46.74 & 52.16 \\
\ours w/o Attn Prun & 60.14 & 46.98 & 52.75  \\
\rowcolor{gray!20} \ours & 62.10 & 46.84 & 53.40 \\
\bottomrule
\end{tabular}
}
\caption{Ablation study on the category supervised visual attention pruning (\%).}
\label{tab:abla_attn_prun_full}
\end{table}

\begin{table}[H]
\centering
\small
\resizebox{\columnwidth}{!}{%
\begin{tabular}{cccc}
\toprule
\multirow{2.5}{*}{Method} &\multicolumn{3}{c}{\bf Single Value Dataset: Pattern} \\
\cmidrule(lr){2-4} 
& {Precision} & {Recall} & {$\text{F}_1$} \\
\midrule
\ours w/o NR & 59.92 & 44.92 & 51.35 \\
\ours w/o Vis-NR & 61.59 & 46.24 & 52.82 \\
\ours w/o Pred-NR  & 60.77 & 45.11 & 51.78 \\
\rowcolor{gray!20} \ours & 62.10 & 46.84 & 53.40 \\
\bottomrule
\end{tabular}
}
\caption{Ablations on the two-level neighborhood-regularized sample weight adjustment (\%).}
\label{tab:abla_neigh_reg_full}
\end{table}

\section{Retrieval Ablation on Pattern Dataset}
\begin{figure}
    \centering
    \includegraphics[width=\linewidth]{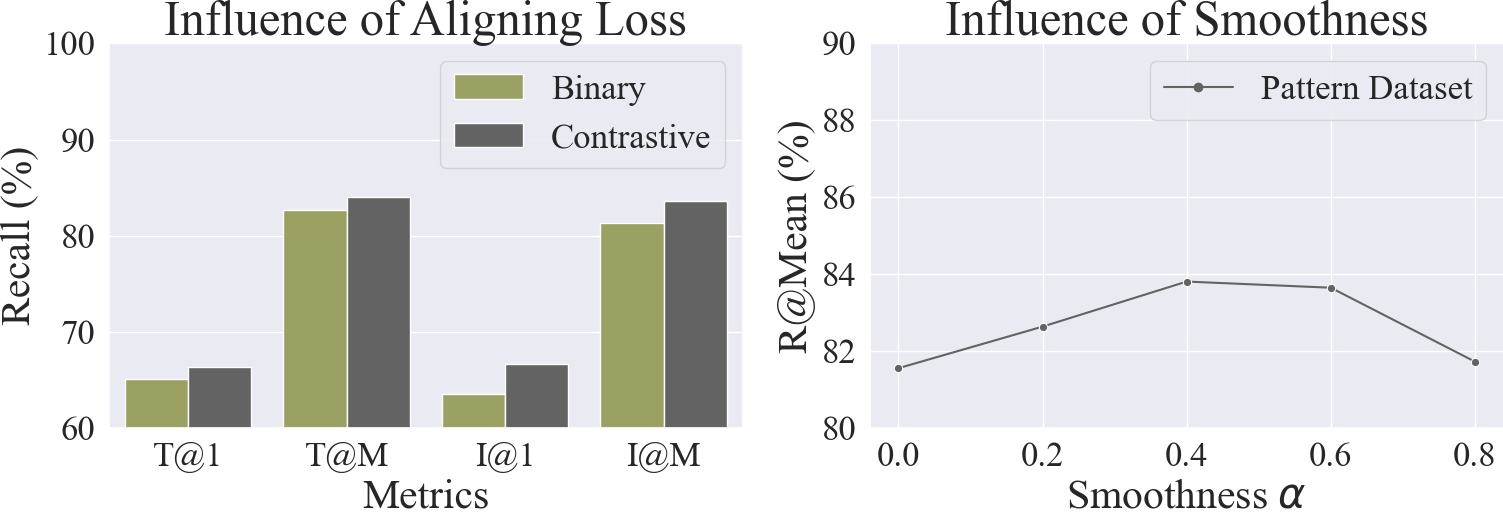}
    \caption{The influence study of alignment objectives, i.e., binary matching v.s. contrastive, and softness $\alpha$ study via cross-modality retrieval on the Pattern dataset. }
    \label{fig:retrieval_pattern}
\end{figure}
Similar to Figure \ref{fig:retrieval}, we also demonstrate the cross-modality retrieval results on the pattern dataset in Figure \ref{fig:retrieval_pattern}. The conclusion is consistent with our observations mentioned in Section \ref{sec:abla-sc}, where the contrastive objective demonstrates advantages in cross-modal alignment and fusion, and the best smoothness choice peaks at 0.4. 

\section{Visualizations of Attention Pruning}
Examples of visualization on the learned attention mask are demonstrated in Figure \ref{fig:attn_vis_appendix}. It is observed that the visual foreground is highlighted under the supervision of category classification, which potentially encourages a higher prediction precision for fine-grained tasks like attribute extraction, as proved by the experimental results.  

\begin{table*}[htbp]
\centering
\small 
\resizebox{\textwidth}{!}{
\begin{tabular}{ll}
\toprule
\bf Category (Product Type) & \bf Candidate Attribute Values Given the Category \\
\midrule
\textcolor{black}{cereal} & grain, flake, seed, liquid, powder, ground \\
\textcolor{black}{dishwasher detergent} & gel, capsule, pac, liquid, tablet, pod, powder \\
\textcolor{black}{face shaping makeup} & powder, pencil, cream, liquid, stick, oil, spray, gel, cushion, blush, drop, balm, gloss \\
\textcolor{black}{fish} & fillet, chunk, steak, solid, stick, whole, slice, ground \\
\textcolor{black}{herb} & powder, root, leaf, thread, flake, seed, tea bag, stick, oil, slice, pod, ground, bean, paste \\
\textcolor{black}{honey} & jelly, capsule, lozenge, candy, cream, powder, granule, flake, liquid, stick, oil, crystal, butter, drop, syrup, comb \\
\textcolor{black}{insect repellent} & wipe, spray, band, granular, liquid, stick, candle, coil, oil, lotion, gel, capsule, tablet, powder, balm, patch, roll on \\
\textcolor{black}{jerky} & strip, slab, shredded, bite, bar, slice, stick, ground \\
\textcolor{black}{sauce} & puree, jelly, paste, seed, liquid, gravy, ground, oil, powder, cream \\
\textcolor{black}{skin cleaning agent} & powder, capsule, toothpaste, wipe, cream, spray, mousse, bar, flake, liquid, lotion, gel, serum, mask, ground, balm, paste, foam \\
\textcolor{black}{skin foundation concealer} & powder, pencil, cream, mousse, liquid, stick, oil, lotion, spray, cushion, gel, drop, serum, balm, airbrush \\
\textcolor{black}{sugar} & granule, crystal, pearl, liquid, powder, cube, ground \\
\textcolor{black}{sunscreen} & wipe, cream, spray, mousse, liquid, ointment, stick, fluid, oil, lotion, milk, compact, gel, drop, serum, powder, balm, foam, mist \\
\textcolor{black}{tea} & leaf, powder, granule, tea bag, liquid, pod, ground, brick \\
\bottomrule
\end{tabular}
}
\caption{The annotation candidates provided to annotators given each sample type on the Item Form dataset.}
\label{tab:annotation-itemform}
\end{table*}

\section{Human Annotation Instruction}
\label{sec:annotation}
We create source-aware fine-grained datasets with internal human annotators. Below are the instruction texts provided to annotators: 

The annotated attribute values are used for research model development of multimodal attribute information extraction and fine-grained error analysis. The datasets are named source-aware multimodal attribute extraction evaluation benchmarks and will be released to facilitate public testing and future studies in bias-reduced multimodal attribute value extraction model designs. All the given sample profiles (title, bullets, and descriptions) and images are collected from the public \verb|amazon.com| web pages, so there is no potential legal or ethical risk for annotators. Specifically, the annotation requirements compose two tasks in order: (1) Firstly, for each given sample\_id in the given ASINs set, first determine the category of the sample by referring to ID2Category.csv mapping file, then label the gold value for the queried attribute by selecting from the candidates given the category. The annotation answer candidates for the Item Form dataset can be referred to in Table \ref{tab:annotation-itemform}. Note that this gold value annotation step requires reference to both sample textual titles, descriptions, and images; (2) For each annotated ASIN, mark down which modality implies the gold value with an additional source label, with different meanings as below: 
\begin{itemize}[nosep,leftmargin=*]
\item \textbf{0}: \textit{the gold attribute value can be found in text.}
\item \textbf{1}: \textit{the gold attribute value cannot be inferred from the text but can be found in the image.}
\end{itemize}
The annotated attribute values and source labels are assembled in fine-grained source-aware evaluation. 

\section{Neighborhood Regularization Demos}
\label{sec:demo}
\begin{figure}
    \centering
    \includegraphics[width=\linewidth]{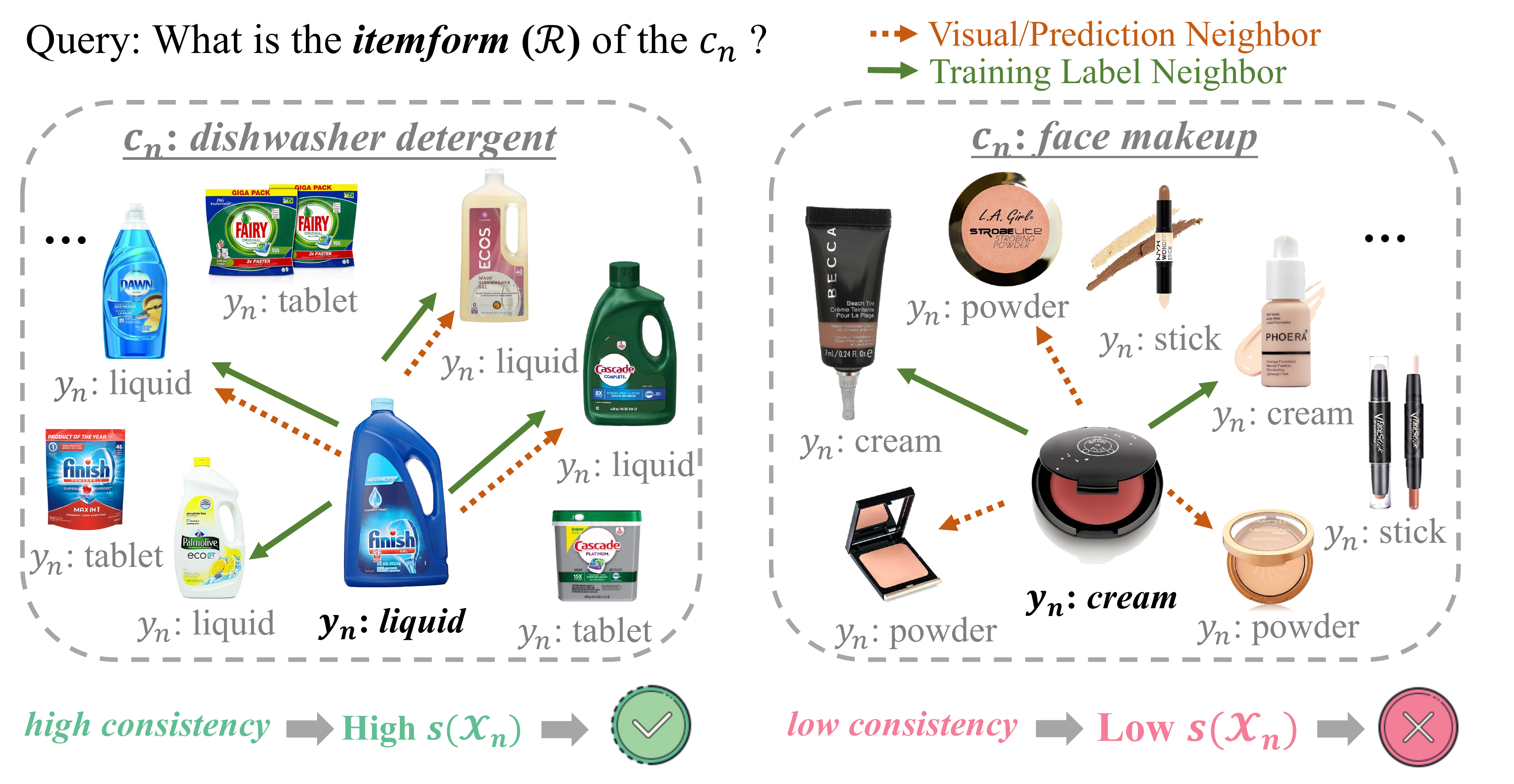}
    \caption{Demo examples for illustrating S3: two-level neighborhood-regularized sample weight adjustment. }
    \label{fig:demo-knn}
\end{figure}

We provide two more demo examples for illustrating the two-level neighborhood-regularized sample weight adjustment in Figure \ref{fig:demo-knn}. The example on the left demonstrates a higher consistency between the green arrows (which point to samples with the same training label as $y_n$) and red arrows (which point to $k$-nearest neighbor samples in visual feature and previous prediction space), indicating a higher reliability of $y_n$. Thus the sample weight of $\mathcal{X}_n$ will be increased in the next training epoch. In contrast, the training label neighbors and visual/prediction neighbors of the right example show a large inconsistency, which implies a relatively lower reliability of $y_n$. Therefore, the sample weight $s\left(\mathcal{X}_n\right)$ of the right $\mathcal{X}_n$ will be degraded in the next epoch. This regularization process adjusts the sample weights of all the training samples in each epoch.

\balance


\end{document}